\documentclass[letterpaper,10pt,conference]{ieeeconf}
\usepackage{etoolbox}
\usepackage{graphicx}
\usepackage{arydshln}
\usepackage{amsmath}
\usepackage{amssymb}
\usepackage{enumerate}
\usepackage{booktabs}
\usepackage{authblk}
\usepackage{color}
\usepackage{url}
\usepackage{multirow}
\usepackage[table]{xcolor}
\usepackage{stfloats}
\usepackage{multirow}
\usepackage{booktabs}
\usepackage{microtype}
\usepackage[pagebackref=true,breaklinks=true,letterpaper=true,colorlinks,bookmarks=false]{hyperref}
\let\labelindent\relax
\usepackage{enumitem}
\setlist[itemize]{leftmargin=*}

\usepackage[font=footnotesize]{caption}

\IEEEoverridecommandlockouts
\begin{document}
\title{CoralVOS: Dataset and Benchmark for Coral Video Segmentation}
\author{Ziqiang~Zheng$^{1*}$,~Yaofeng~Xie$^{2}$,~Haixin~Liang$^{3}$,~Zhibin~Yu$^{2}$,~Sai-Kit~Yeung$^{1,3}$
 \thanks{$^1$Ziqiang Zheng and Sai-Kit Yeung are with the Department of Computer Science and Engineering, The Hong Kong University of Science and Technology. 
 }
 \thanks{$^2$Yaofeng Xie and Zhibin Yu are with the School of Electronic Information Engineering/the Key Laboratory of Ocean Observation and Information of Hainan Province, Faculty of Information Science and Engineering/Sanya Oceanographic Institution, Ocean University of China, Qingdao/Sanya, China.}
 \thanks{$^3$Haixin Liang and Sai-Kit Yeung are with the Division of Integrative Systems and Design, Hong Kong University of Science and Technology.}
 \thanks{$^*$Corresponding author: Ziqiang Zheng (zhengziqiang1@gmail.com)}
\vspace{-3em}
}

\let\oldtwocolumn\twocolumn
\renewcommand\twocolumn[1][]{%
    \oldtwocolumn[{#1}{
    \begin{center}
            \vspace{-1.5em}
           \includegraphics[width=\textwidth]{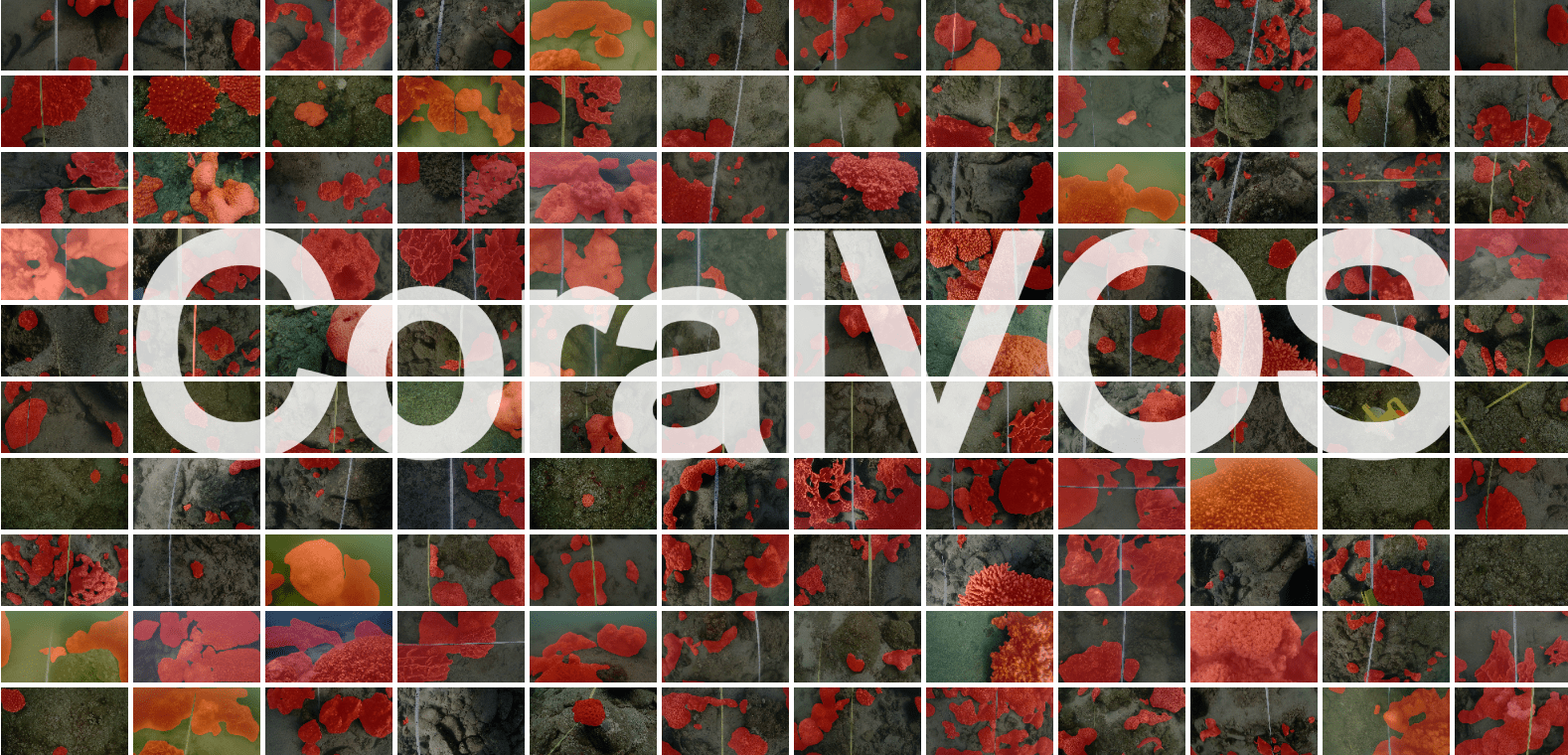}
           \captionof{figure}{Example images with mask annotations from our CoralVOS dataset. The CoralVOS dataset could support segmenting different types of corals.}
           \label{fig:teaser}
           \vspace{-0.5em}
        \end{center}
    }]
}

\maketitle
\makeatletter
\patchcmd{\@makecaption}
  {\scshape}
  {}
  {}
  {}
\makeatother

\begin{abstract}
Coral reefs formulate the most valuable and productive marine ecosystems, providing habitat for many marine species. Coral reef surveying and analysis are currently confined to coral experts who invest substantial effort in generating comprehensive and dependable reports (\emph{e.g.}, coral coverage, population, spatial distribution, \textit{etc}), from the collected survey data. However, performing dense coral analysis based on manual efforts is significantly time-consuming, the existing coral analysis algorithms compromise and opt for performing down-sampling and only conducting sparse point-based coral analysis within selected frames. However, such down-sampling will \textbf{inevitable} introduce the estimation bias or even lead to wrong results. To address this issue, we propose to perform \textbf{dense coral video segmentation}, with no down-sampling involved. Through video object segmentation, we could generate more \textit{reliable} and \textit{in-depth} coral analysis than the existing coral reef analysis algorithms. To boost such dense coral analysis, we propose a large-scale coral video segmentation dataset: \textbf{CoralVOS} as demonstrated in Fig.~\ref{fig:teaser}. To the best of our knowledge, our CoralVOS is the first dataset and benchmark supporting dense coral video segmentation. We perform experiments on our CoralVOS dataset, including 6 recent state-of-the-art video object segmentation (VOS) algorithms. We fine-tuned these VOS algorithms on our CoralVOS dataset and achieved observable performance improvement. The results show that there is still great potential for further promoting the segmentation accuracy. The dataset and trained models will be released with the acceptance of this work to foster the coral reef research community.
\end{abstract}

\section{Introduction}
Coral reefs represent one of the planet's most diverse and productive ecosystems, providing habitat and shelter for a vast range of marine species. Performing underwater coral reef monitoring~\cite{edwards2017large, MUSCD, levy2022emerging, beijbom2012automated,ahmad2020comparison,cinner2016bright,haas2016global} can identify and track changes in coral reef health, understand the impacts of human activities on coral reefs, and help maintain the coral biological diversity. With more advanced autonomous underwater vehicles (AUVs)~\cite{cho2017auv} and remotely operated underwater vehicles (ROVs)~\cite{huvenne2018rovs} deployed, the acquisition of underwater coral reef images/videos becomes more convenient and efficient, resulting in a large number of underwater videos collected for different purposes.

With the significant amount of coral surveying videos, coral reef video analysis has gained increasing attention. Coral video analysis~\cite{modasshir2020enhancing} allows researchers to analyze video footage of coral reefs and track changes in coral cover and health over time. This helps in monitoring the condition and dynamics of coral reefs, assessing the impact of environmental stressors on coral coverage, and identifying areas in need of conservation efforts. It also enables the quantification of the coral population of different sites and countries~\cite{cinner2016bright}. This information aids in understanding the overall distribution of coral communities and provides insights into conversational efforts and policy-making. Among video analysis, video object segmentation (VOS) is the most useful and effective way for dense coral video analysis. 

\begin{figure}[t]
  \begin{center}
    \includegraphics[width=\linewidth]{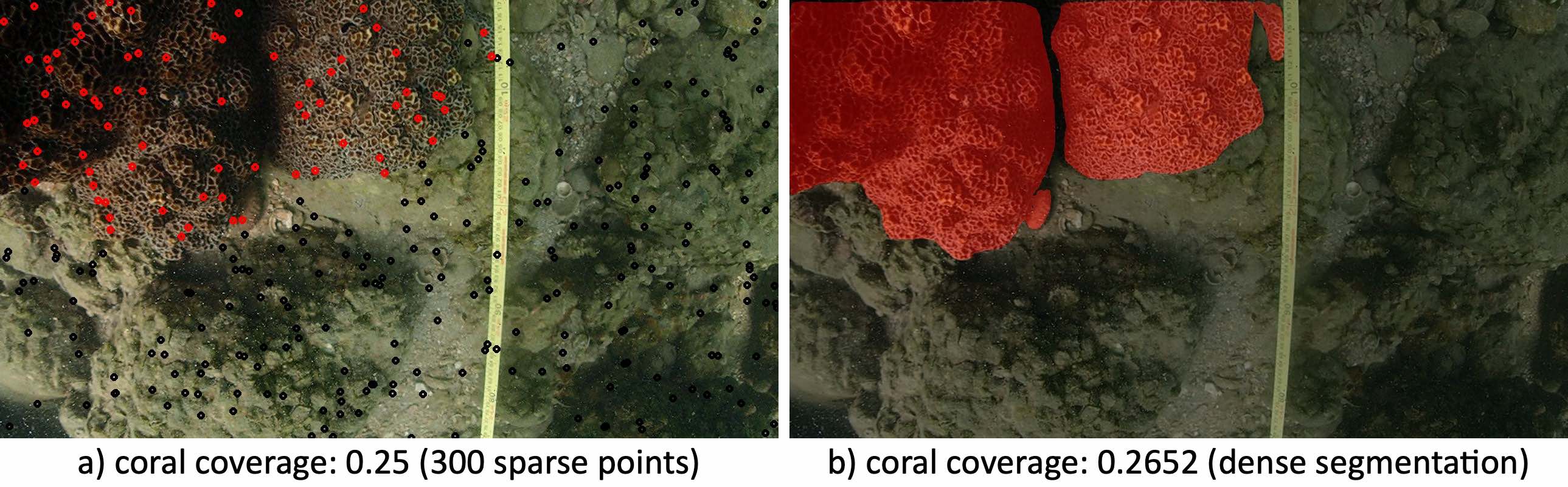}
  \end{center}
  \caption{Comparison between computing coral coverage from sparse point (300 sparse points) based analysis and dense coral segmentation.}
  \label{fig:sparse_dense}

\end{figure}

Different from the existing dominant coral analysis algorithms~\cite{kohler2006coral,pavoni2022taglab,king2018comparison}, which usually pick up some frames from the whole video sequence for down-sampled sparse point based coral analysis~\cite{carleton1995quantitative}, in this work, we propose to perform \textbf{dense coral segmentation}. We argue that the down-sampling involved in the existing coral analysis algorithms~\cite{tabugo2016coral,pavoni2022taglab} will inevitably introduce bias to the estimation results and tend to miss some key information or even lead to some inaccurate estimation results~\cite{jokiel2015comparison} compared with dense segmentation as demonstrated in Fig.~\ref{fig:sparse_dense}. 

Understanding the coral reef ecosystem (including detailed coral distribution and coral coverage) should be delineated based on video data. On one hand, coral reef videos contain richer information (\emph{e.g.}, motion pattern of different objects and temporal consistency) than the single coral reef image, and thus provide more cues for coral analysis. On the other hand, the coral reef video analysis supports more \textbf{reliable}, \textbf{stable} and \textbf{denser} statistics analysis without any down-sampling involved, yielding a more comprehensive coral reef report (\emph{e.g.}, cover percentage, population, and spatial distribution). 

\begin{figure}[t]
  \begin{center}
    \includegraphics[width=\linewidth]{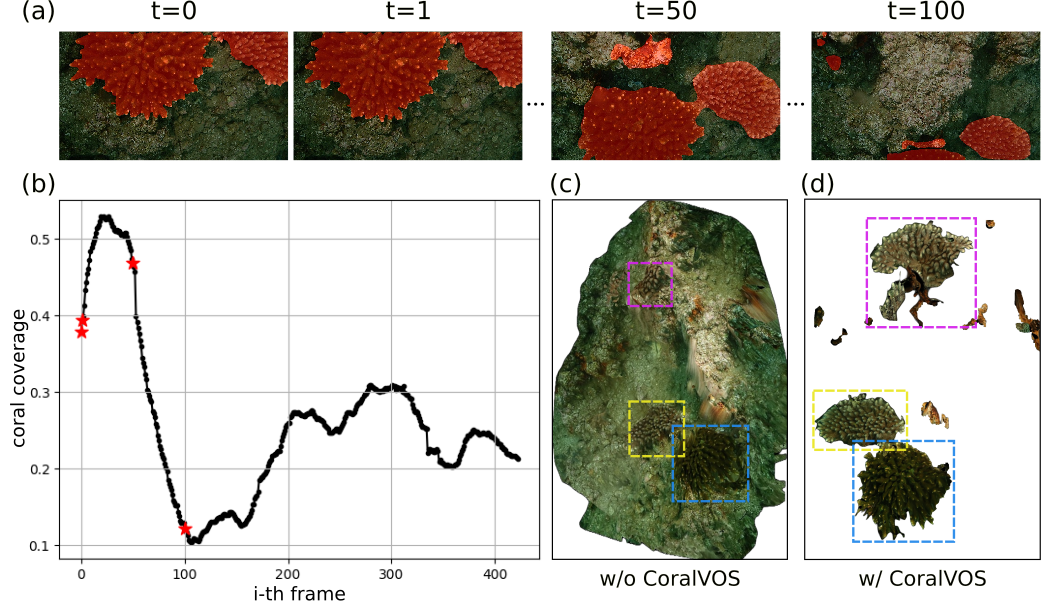}
  \end{center}
  \caption{Dense coral video segmentation could support more reliable and in-depth coral analysis in a), yielding the coral coverage, population, and spatial distribution in b). It also leads to better 3D coral reconstruction in d) compared with the setting without dense coral video segmentation in c).}
  \label{fig:demo}
\end{figure}

In this work, we perform \textbf{dense coral video segmentation}, which indicates \textbf{all the pixels within each frame} from the coral reef video sequences have been considered during the analysis procedure. Besides, we can also better monitor the spatial coral distribution from the coral coverage curve and support better 3D coral reconstruction (removing the non-coral background and alleviating geometry distortions) as illustrated in Fig.~\ref{fig:demo}. These valuable information are crucial for coral reef monitoring~\cite{cinner2016bright}, marine spatial planning~\cite{levy2022emerging}, and conservation prioritization~\cite{haas2016global,darling2019social}. Through dense coral video segmentation, we could assess the suitability of areas for coral restoration efforts. By understanding the distribution of existing coral colonies, researchers can identify potential sites for successful restoration projects.

Despite the overwhelming advantage of dense coral video segmentation, we notice there are relatively few or no research works that focus on dense coral video segmentation. The existing coral datasets~\cite{beijbom2016improving} mainly focus on image-level analysis, only utilizing the down-sampled images from the whole video for analysis. One potential reason that coral video segmentation is less explored, may come from the lack of a large-scale dataset for fully supervised training. The appearance and motion of coral objects can change significantly in video frames, which also makes it difficult to segment corals accurately. In this work, we propose the first large-scale dense coral video segmentation dataset named \textbf{CoralVOS}, which is collected from 17 different sites and with 150 videos (60,456 densely labeled frames in total) for supervised training and evaluation. We also notice most existing VOS datasets~\cite{perazzi2016benchmark,pont20172017,vos2018} usually assume that the camera is static while the interest of objects is moving, or the camera and the object are in relative motion to ensure that the entire object can be fully encompassed. Differently, the coral reef surveying video sequence is captured in a fully contrast way due to its intrinsic requirement~\cite{safuan2015optimization}: the coral is static while the camera is constantly moving following the transect line~\cite{ahmad2020comparison}, resulting in uncertainty and noise when segmenting the new coming frames. Our CoralVOS could heavily promote the development of coral surveying analysis. The main contributions of this paper are as follows:
\begin{itemize}
    \item A large-scale coral video segmentation benchmark to boost learning-based coral reef surveying and analysis. Our CoralVOS dataset has a large range of illumination, appearance, complexity, and visibility changes.
    \item We have benchmarked six existing state-of-the-art VOS algorithms on the proposed CoralVOS dataset. We observe that there is still a large room for promoting the dense coral video segmentation performance.
    \item To the best of our knowledge, CoralVOS is the first dense coral video segmentation dataset and benchmark for coral analysis. We demonstrate that CoralVOS could significantly promote coral population estimation, spatial coral reef modeling, and 3D coral reef reconstruction. 
\end{itemize}

\section{Related Work}
\subsection{Coral Surveying and Analysis}
The methods of monitoring and surveying coral reefs encompass the use of scuba divers~\cite{safuan2015optimization} and autonomous or remotely operated vehicles~\cite{cho2017auv,huvenne2018rovs,rodriguez2014vision,sadrfaridpour2021detecting}. With collected coral reef surveying images/videos, to achieve effective and efficient coral analysis, various coral reef labeling and analysis tools, including Coral Point Count with Excel Extensions (CPCe)~\cite{kohler2006coral}, PhotoQuad~\cite{trygonis2012photoquad}, BIIGLE~\cite{langenkamper2017biigle} and CoralNet~\cite{beijbom2015towards} have been developed. Most existing tools only support annotating sparse points or bounding boxes, which cannot provide a dense analysis of the coral reefs. The coral experts then analyze the annotated data to determine the species~\cite{beijbom2012automated}, health~\cite{neal2017caribbean}, and population diversity~\cite{zheng2023marine} of the coral reefs. However, the whole analysis procedure is tedious and time-consuming. Besides, the existing coral research is mainly limited to the images while not considering the whole video sequence as input. ~\cite{modasshir2020enhancing} first proposed to conduct the coral reef localization for the whole coral reef video. However, we cannot densely compute precise and accurate coral coverage and population based on the detected bounding boxes since the corals usually have irregular boundaries. In this work, we aim to push the boundaries of coral reef understanding to video analysis and pave the way for dense coral video segmentation. 

\subsection{Video Object Segmentation}
VOS is a fundamental and challenging problem in computer vision and robotics fields, with numerous potential applications including autonomous driving~\cite{nagai2022high,botach2022end,siam2019video,behl2020meta}, robotics~\cite{walther2004detection,wang2019hand}, automated surveillance~\cite{patil2021unified}, underwater exploring~\cite{zhang2022coastal,modasshir2020enhancing}, and video conferencing~\cite{ackermann2023maskomaly,botach2022end}. VOS~\cite{cheng2021rethinking,yang2021associating,cheng2021modular,cheng2022xmem} aims to propagate the given mask of the initial frame to other consecutive frames of the video sequence,  where image pixels are densely predicted. The visual similarity between frames, motion cues, and temporal consistency among the whole video are utilized for identifying the same object across the video. The designed algorithms~\cite{yang2021associating,yang2021collaborative,cheng2021modular} are supposed to consider the target objects as general objects and do \textbf{not} care about the semantics. Besides VOS, the recent Segment Anything model~\cite{kirillov2023segment} (SAM) has demonstrated an efficient zero-shot ability to yield precise masks for unseen object categories. Based on SAM, SAM-Track~\cite{cheng2023segment} employs multi-modal interactions that enable users to select multiple objects in videos for tracking~\cite{yang2023track} and segmenting objects in videos in an interactive way while not requiring the recognition of the object categories. This work aims to provide a large-scale coral video segmentation dataset through an interactive labeling tool. We also demonstrate the essential differences between performing VOS for coral reef analysis (\textit{domain-specific}) and in-air general-purpose VOS.

\section{CoralVOS}
\subsection{Problem Formulation}
\textit{Coral video object segmentation} is a binary labeling problem aiming to separate foreground object(s) from the background region of a video. Given a sequence of video frames, denoted as $\{I_t\}_{t=1}^T$, where $T$ is the total number of frames, the goal of video object segmentation is to assign binary labels to each pixel in each frame, indicating whether it belongs to the object of interest (foreground) or the background. Formally, for each frame $I_t$, we seek to find a binary mask $M_t$, where $M_t(i,j) = 1$ if pixel $(i,j)$ belongs to the object and $M_t(i,j) = 0$ if it belongs to the background. Notably, only the binary mask $M_1$ of the first frame is provided as an initial reference. The challenge lies in accurately and consistently segmenting the object across all frames, accounting for variations in object appearance, shape, and motion, as well as handling occlusions and complex background scenarios. The objective is to develop an effective video object segmentation algorithm that produces accurate binary masks $M_t$ for each frame, starting from the initial mask $M_1$, enabling the precise delineation of the object of interest in the video sequence.

\subsection{CoralVOS Dataset}
We have collected 150 video sequences for dense coral video segmentation from 17 different sites: 100 for training, 25 video sequences for validation, and the remaining 25 sequences withheld for testing. All these video sequences are collected with the benthic view. We collect the videos under challenging and in-the-wild conditions (\emph{e.g.}, with low visibility, background clutter, motion blur, occlusion, dynamic illumination, color distortion, and optical artifacts). The FPS is set to 25 and the image resolution is $1280\times 720$. Each video sequence lasts at least 236 frames. In total, 60,456 frames are densely labeled. 

\noindent\textbf{Video labeling tool}. We have developed an interactive coral labeling tool to reduce the labeling time and promote the labeling efficiency of coral video labeling. The SAM model~\cite{kirillov2023segment} is integrated to generate accurate and precise coral masks based on user point prompts. Then, we adopt the XMem~\cite{cheng2022xmem} to propagate the labeled coral mask to the consecutive frames. When the users feel that the propagated masks are not accurate enough, the experts could remove or refine the propagated coral masks to obtain more consistent and accurate labels. The refined coral mask will be overwritten into the system for label propagation. 

\noindent\textbf{Labeling rule}. We follow the labeling rules for performing binary coral discrimination. We only label coral masks when clear and visible to discriminate the coral instances (closer, less blurry, and well-exposed) from the background. Due to poor visibility of the specific underwater conditions, objects more than a few meters away are difficult for even coral experts to recognize. Thus, in this work, we do \textbf{not} consider to provide the coral species annotations. Instead, we only perform the binary coral labeling while ignoring the species-level coral annotations.

\begin{table*}
    \caption{}
    \vspace{-0.07in}
    \caption*{Direct comparison between DAVIS-2016~\cite{perazzi2016benchmark}, DAVIS-2017~\cite{pont20172017}, YouTube-VOS~\cite{vos2018}, and our proposed CoralVOS dataset according to different properties.\label{table:comp}}
	\centering
\resizebox{0.95\linewidth}{!}{
\begin{tabular}{c|c|c|c|c|c|c|c|c}
    \toprule
    Datasets  & Sequences & Images & Duration (min) & Purpose & Diversity & Turbidity & Motion Blur & Complexity 
    \\ \midrule 
     DAVIS-2016~\cite{perazzi2016benchmark} & 50 & 3,400 & 2.28 & General-purpose  & Low & Clean & $\times$ & Low \\
     DAVIS-2017~\cite{pont20172017} & 90 & 10,731 & 5.17 & General-purpose & Medium & Clean & $\times$ & Low \\
     YouTube-VOS~\cite{vos2018}  & 4,453 &  197,272 & 334.81 & General-purpose & High & Clean & $\times$ & High \\
     \midrule
     \rowcolor[gray]{0.84}CoralVOS  & 150 & 60,456 & 48.17 & Domain-specific  & Medium & Turbid & $\checkmark$ & High  \\
    \bottomrule
    \end{tabular}
}
\end{table*}

\subsection{Comparison with Previous Benchmarks}
\begin{figure}[t]
  \centering
  \includegraphics[width=\linewidth]{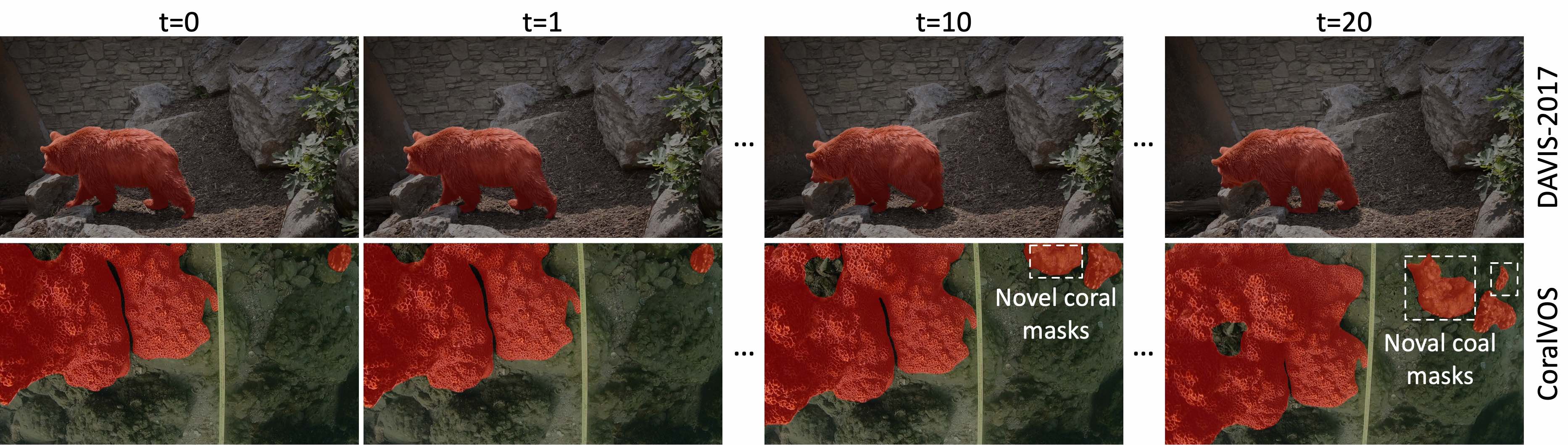}
  \caption{A direct comparison between the video sequence from DAVIS-2017 dataset~\cite{pont20172017} and our CoralVOS dataset. } 
  \label{fig:problem}
\end{figure}
We compare the proposed \textbf{CoralVOS} with the existing benchmarks from two aspects: 1) \textit{video object segmentation} and 2) \textit{coral analysis}. 

\noindent\textbf{Video object segmentation}. To evaluate and boost the performance of VOS algorithms, some widely used video object segmentation datasets have been proposed:
\begin{itemize}
    \item \textbf{DAVIS-2016}~\cite{perazzi2016benchmark} is a dataset for VOS which consists of 50 videos in total (30 videos for training and 20 for testing). Per-frame pixel-wise annotations are offered.
    \item  \textbf{DAVIS-2017}~\cite{pont20172017} contains 150 high-resolution videos collected and 94 common object categories. The length of each video is around 3 to 6 seconds.
    \item \textbf{YouTube-VOS}~\cite{vos2018} is a large-scale dataset, including the training set (3,471 videos), validation set (507 videos), and testing set (541 videos). Instance-level annotations are provided every 5 frames in a 30FPS frame rate. 
\end{itemize}
We provide a direct comparison between these datasets and our CoralVOS dataset in Table~\ref{table:comp}. Compared with the existing video segmentation datasets, our CoralVOS has such essential differences as demonstrated in Fig.~\ref{fig:problem}. Our CoralVOS serves for coral surveying and monitoring, in which the camera is constantly moving following the transect line~\cite{jokiel2015comparison,safuan2015optimization} while corals remain static. In contrast, the existing VOS dataset~\cite{perazzi2016benchmark,pont20172017} usually assumes that the camera is static while the object is moving or that the camera and objects are in relative motion. More importantly, different from the existing VOS datasets, in which the \textbf{holistic view} of the object is given for propagating the mask of the initial frame to consecutive frames, there are always \textbf{novel coral masks} appearing due to the camera is constantly moving. Such special attribute of the surveying videos introduces \textbf{uncertainty} and \textbf{noise} when propagating the mask of previous frames to new coming frames. 

\begin{table}
     \caption{}
     \vspace{-0.07in}
    \caption*{Direct comparison between Eilat~\cite{beijbom2016improving}, CoralNet~\cite{beijbom2015towards}, Mosaics UCSD~\cite{edwards2017large} and our CoralVOS.\label{table:coral_comp}}
	\centering
\resizebox{\linewidth}{!}{
\begin{tabular}{c|c|c|c|c|c}
    \toprule
    Datasets  & Images & Purpose & VOS  & Turbidity & Motion Blur  
    \\ \midrule 
     Eilat~\cite{beijbom2016improving} & 142 & Classification & $\times$ & Clean & $\times$ \\
     CoralNet~\cite{beijbom2015towards} & 416,512  & Sparsely annotated & $\times$  &  Clean & $\times$ \\
     Mosaics UCSD~\cite{edwards2017large} & 4,193 & Dense segmentation & $\times$ &  Clean & $\times$  \\\midrule
     \rowcolor[gray]{0.84}CoralVOS & 60,456 & Dense segmentation & $\checkmark$ & Turbid & $\checkmark$ \\
    \bottomrule
    \end{tabular}
}
\end{table}

\noindent\textbf{Coral analysis}. Similarly, we summarize the recent coral reef datasets as follows:
\begin{itemize}
    \item \textbf{Eilat Fluorescence} dataset~\cite{beijbom2016improving} consists of 142 training images and 70 test images. All images are with 200 sparse point labels arranged as a grid in the center of each image. 
    \item \textbf{Mosaics UCSD}~\cite{edwards2017large} is the only publicly available dataset, which supports \textbf{dense coral genus segmentation} with ground truth masks. It contains 4,193 training images and 729 test images with 34 semantic classes. 
    \item \textbf{CoralNet}~\cite{beijbom2015towards} dataset is a large-scale coral reef surveying dataset, providing the sparse point annotations. CoralNet contains 416,512 images taken across years with approximately 400,000 manually annotated sparse point labels.
\end{itemize}
We also directly review existing coral datasets in Table~\ref{table:coral_comp}. Unlike existing coral reef datasets, we propose the first \textbf{dense coral video object segmentation dataset} to support comprehensive and in-depth coral analysis of coral reef surveying in the wild. All the videos of our CoralVOS dataset are captured by scuba divers who have specific expertise when collecting the coral surveying videos, or the AUVs/ROVs following some pre-defined transect lines.

\subsection{Challenges of CoralVOS}
There are some challenging scenarios (\emph{e.g.}, \textit{low visibility}, \textit{background clutter}, \textit{motion blur}, \textit{occlusion}, \textit{dynamic lighting}, \textit{color distortion} and \textit{optical artifacts}) in our CoralVOS dataset. We summarize the challenges as follows: 1) The appearance and motion of objects can change \textbf{significantly} in video frames, making it difficult to segment them accurately. 2) Corals can also exhibit different \textbf{deformations}, \textbf{rotations}, and \textbf{scaling} in different frames. The model is required to output consistent predictions. 3) \textbf{Occlusion} occurs when the corals are partially or completely hidden by other objects or the background. 4) \textbf{Motion blur} can cause the background to be chaotic and turbulent, leading to false positives and tracking errors. 5) \textbf{Illumination changes} and \textbf{dynamic lighting} can affect the appearance of coral objects. Such complicated challenges lead to performing dense coral video segmentation still remains an intricate problem.

\section{Experiments}
\subsection{Implementation Details and Evaluation Metrics}
\noindent\textbf{Implementation details}. We have benchmark six existing state-of-the-art video segmentation algorithms on proposed CoralVOS dataset: including AOT~\cite{yang2021associating}, STCN~\cite{cheng2021rethinking}, MiVOS~\cite{cheng2021modular}, DeAOT~\cite{yang2022decoupling}, XMem~\cite{cheng2022xmem} and DEVA~\cite{cheng2023tracking}. Furthermore, we also adopt the SegFormer~\cite{xie2021segformer} to perform the frame-by-frame segmentation. For VOS algorithms, we conduct experiments under two settings: without fine-tuning and with fine-tuning on our CoralVOS dataset. Under the former setting, we adopt the released pre-trained models on DAVIS-2017 and YouTube-VOS datasets for inference. Under the second ``fine-tuning'' setting, we have fine-tuned the pre-trained model to the coral reef field based on the training set of our CoralVOS dataset. We compute quantitative results on the validation set under both settings. For SegFormer, we conduct experiments under the same train/val data split. All the labeled frames from the training set are used for training. Then, we segment video sequences from the validation set frame-by-frame. We perform all the experiments under the default setting for a fair comparison. As for DEVA~\cite{cheng2023tracking}, which is built on GroundingDINO~\cite{liu2023grounding}, we did not fine-tune the pre-trained model since the training codes of GroundingDINO are not released.

\noindent\textbf{Evaluation Metrics}. Evaluating coral video segmentation involves a comprehensive analysis utilizing a range of meticulously designed metrics for assessing the accuracy of VOS algorithms. Following the evaluation metrics of existing benchmarks~\cite{perazzi2016benchmark,pont20172017}, we compute the region similarity $\mathcal{J}$ and the contour accuracy $\mathcal{F}$. Given the segmentation predictions $\hat{M}\in \{0,1\}^{H\times W}$ and our manually labeled ground truth $M \in \{0,1\}^{H\times W}$, where $H$ and $W$ indicate the height and width of the images. We compute $\mathcal{J}$ based on calculating the Intersection over Union (\textbf{IoU}) between $\hat{M}$ and $M$,
\begin{equation}
    \mathcal{J}=\frac{\hat{M}\cap M}{\hat{M}\cup M}.
\end{equation}
We calculate average region similarity over all frames as the final region similarity result. To measure the contour quality of $\hat{M}$, we calculate contour recall $R_c$ and precision $P_c$ via bipartite graph matching~\cite{martin2004learning}. The contour accuracy $\mathcal{F}$ is the harmonic mean of the contour recall $R_c$ and precision $P_c$:
\begin{equation}
    \mathcal{F}=\frac{2P_cR_c}{P_c+R_c},
\end{equation}
which represents how closely the contours of predicted masks resemble those of ground-truth masks. The average contour accuracy $\mathcal{F}$ over all the frames is calculated as the final region similarity result. $\mathcal{J}\& \mathcal{F}=(\mathcal{J}+\mathcal{F})/2$ is used to measure the overall performance.

\subsection{Benchmark with SOTAs}
\begin{figure*}[t]
  \begin{center}
  \includegraphics[width=\linewidth]{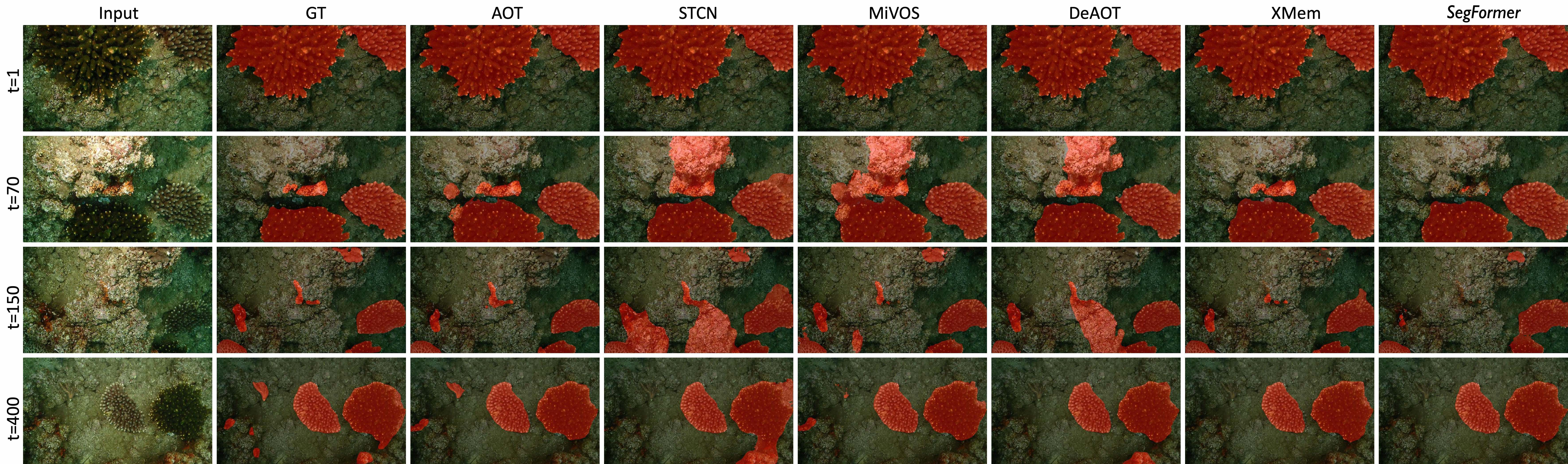}  
  \end{center}
  \caption{The qualitative coral video segmentation result comparison between different algorithms. SegFormer is tested by frame-by-frame segmentation.}
  \label{fig:comp}
\end{figure*}

\begin{table}[!hbt]
    \caption{}
    \vspace{-0.07in}
    \caption*{We report the coral reef video object segmentation results under different settings. The results on DAVIS-2017 dataset are reported for reference. Best results are in \textbf{bold}. $\dagger$ indicates that the model was tested by frame-by-frame.\label{table:result_comp}}
	\centering
\scalebox{0.9}{\begin{tabular}{c|c|ccc|c}
 \toprule
\multirow{2}{*}{Method} & \multirow{2}{*}{\begin{tabular}[c]{@{}c@{}}Fine-tuned\\ on CoralVOS\end{tabular}} & \multicolumn{3}{c|}{CoralVOS} & \multicolumn{1}{c}{DAVIS$_{17}$} \\
& & $\mathcal{J}$  & $\mathcal{F}$ & $\mathcal{J}\& \mathcal{F}$ & $\mathcal{J}\& \mathcal{F}$ \\
\midrule
\multirow{2}{*}{AOT~\cite{yang2021associating}} & $\times$ &  47.40 & 46.17 &  46.79 & 79.2 \\
 & $\checkmark$ & \cellcolor[gray]{0.84}73.36 & \cellcolor[gray]{0.84}68.73 & \cellcolor[gray]{0.84}71.04 &  $-$ \\
\midrule
\multirow{2}{*}{STCN~\cite{cheng2021rethinking}} & $\times$ &  35.32  & 34.11  & 34.72  &  83.0   \\
 & $\checkmark$ &  \cellcolor[gray]{0.84}\textbf{80.32} & \cellcolor[gray]{0.84}\textbf{75.61} & \cellcolor[gray]{0.84}\textbf{77.96}  &  $-$ \\
\midrule
\multirow{2}{*}{MiVOS~\cite{cheng2021modular}} & $\times$ &  39.26 &  35.17  &  37.22 &  84.3   \\
 & $\checkmark$  & \cellcolor[gray]{0.84}78.32 & \cellcolor[gray]{0.84}72.65 & \cellcolor[gray]{0.84}75.49 & $-$ \\
\midrule
\multirow{2}{*}{DeAOT~\cite{yang2022decoupling}} & $\times$ & 37.88 & 38.23 & 38.06 &  85.2 \\
 & $\checkmark$  &  \cellcolor[gray]{0.84}77.21 & \cellcolor[gray]{0.84}74.04 & \cellcolor[gray]{0.84}75.63 &  $-$ \\
\midrule
\multirow{2}{*}{XMem~\cite{cheng2022xmem}}  & $\times$ & 32.68   &  32.21  &  32.44  &  86.2   \\
  & $\checkmark$ & \cellcolor[gray]{0.84}78.11 & \cellcolor[gray]{0.84}74.39 & \cellcolor[gray]{0.84}76.25 &  $-$ \\
\midrule
DEVA~\cite{cheng2023tracking} & $\times$ & 34.99 & 34.81 & 34.90  &  $-$   \\
\midrule
SegFormer~\cite{xie2021segformer}$^{\dagger}$ & $\checkmark$ & 76.87 & 68.87 & 72.87 & $-$ \\
\bottomrule 
\end{tabular}}
\end{table}

We first provide the quantitative results of different algorithms under different settings in Table~\ref{table:result_comp}. As illustrated, directly adopting the pre-trained general-purpose models on the domain-specific coral reef analysis tasks results in poor video segmentation results. While these models have demonstrated satisfactory performance on general objects in typical environments, as evidenced by their results on the DAVIS-2017 dataset, they face inherent challenges in the underwater environment, including the constant emergence of novel coral masks. The models have lost the coral masks for tracking and propagating. We attribute this failure to the essential difference between in-air VOS and our coral VOS designed for underwater coral reef surveying and exploring. Besides, the pre-trained modes cannot well recognize and segment the corals without fine-tuning. Besides, we notice that AOT and DeAOT with more lightweight network backbones demonstrate better coral VOS performance than XMem~\cite{cheng2022xmem} and STCN~\cite{cheng2021rethinking} under the setting without the fine-tuning. The possible reason may be that deeper models may tend to overfit the task of segmenting general objects in typical in-air environments.

After fine-tuning the pre-trained models on our CoralVOS dataset, the ability of various VOS models to recognize the corals has been greatly promoted. Thus, observable performance gain has been achieved as reported in Table~\ref{table:result_comp}. We provide the corresponding qualitative results of different VOS algorithms after the fine-tuning in Fig.~\ref{fig:comp}. Besides, we have also reported the results of segmenting the coral sequence frame-by-frame based on SegFormer in Fig.~\ref{fig:comp} and Table~\ref{table:result_comp}, respectively. We demonstrate that the proposed coral VOS can achieve better and more consistent coral segmentation results than performing coral segmentation frame-by-frame. Finally, we also observe that there is still a great potential for further promoting the coral VOS performance.  

\subsection{Dense Coral Analysis}
\begin{figure}[t]
  \begin{center}
    \includegraphics[width=\linewidth]{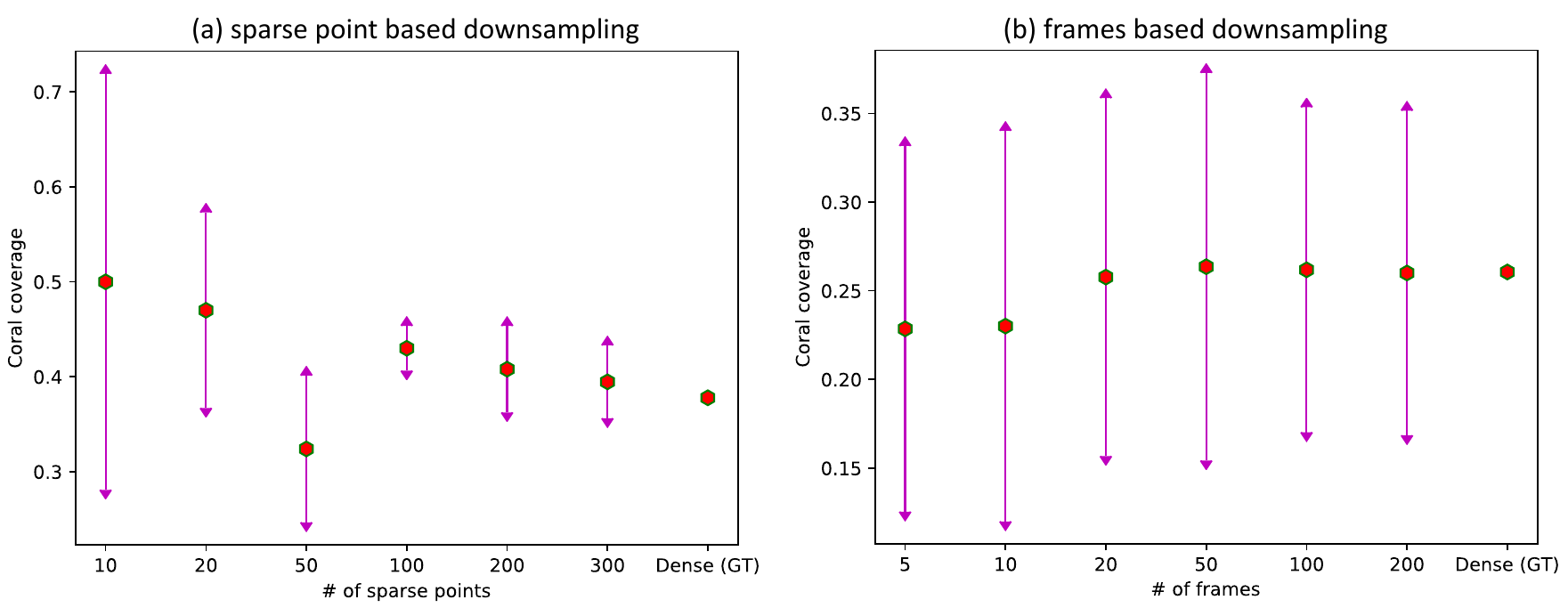}
  \end{center}
  \caption{We present the difference between dense coral reef analysis with the existing sparse point based analysis in a) and frames based coral video analysis in b).}
  \label{fig:dense}
\end{figure}

In this section, we demonstrate that the proposed CoralVOS could significantly promote the stability and efficiency of coral analysis. We first demonstrate the overwhelming advantage of dense coral reef analysis over the existing sparse point based analysis algorithms~\cite{kohler2006coral,tabugo2016coral} in Fig.~\ref{fig:dense} a). We take the first frame video sequence ``No. 102'' as an illustration. We randomly choose 10, 20, 50, 100, 200, and 300 sparse points and compute the corresponding coral coverage results under these settings. We repeat such sampling 5 times under each setting for computing the mean and standard deviation values. We regard our manually labeled dense pixel annotation as ground truth (GT). As illustrated, with the more sparse points sampled, we can obtain more stable and accurate coral coverage estimation results. However, sampling more sparse points usually indicates linearly increasing labeling time. Our dense coral analysis results could be regarded as the upper/optimal bound of coral coverage estimation since it takes all the pixels within the coral images into account.

Besides, given the coral reef surveying video (``No. 102'') with only the first frame labeled, we perform the dense coral video segmentation and compute final average coral coverage based on all video frame as GT. Similarly, we randomly sample different numbers (5, 10, 20, 50, 100 and 200) of frames from the whole video sequence. We directly average the coverage results of these selected frames (300 sparse points are used for computing coverage results for each frame) to obtain the final coral coverage result. We repeat such experiments 5 times to obtain the mean and standard deviation values in Fig.~\ref{fig:dense} b). With more frames sampled, we could obtain more accurate coral coverage result for the whole video sequence. However, we still observe a large standard deviation value and we attribute this to estimation bias caused by the sparse point sampling. In contrast, dense coral video segmentation could obtain more reliable and stable coral coverage estimation results. 

\begin{figure}[t]
  \begin{center}
    \includegraphics[width=\linewidth]{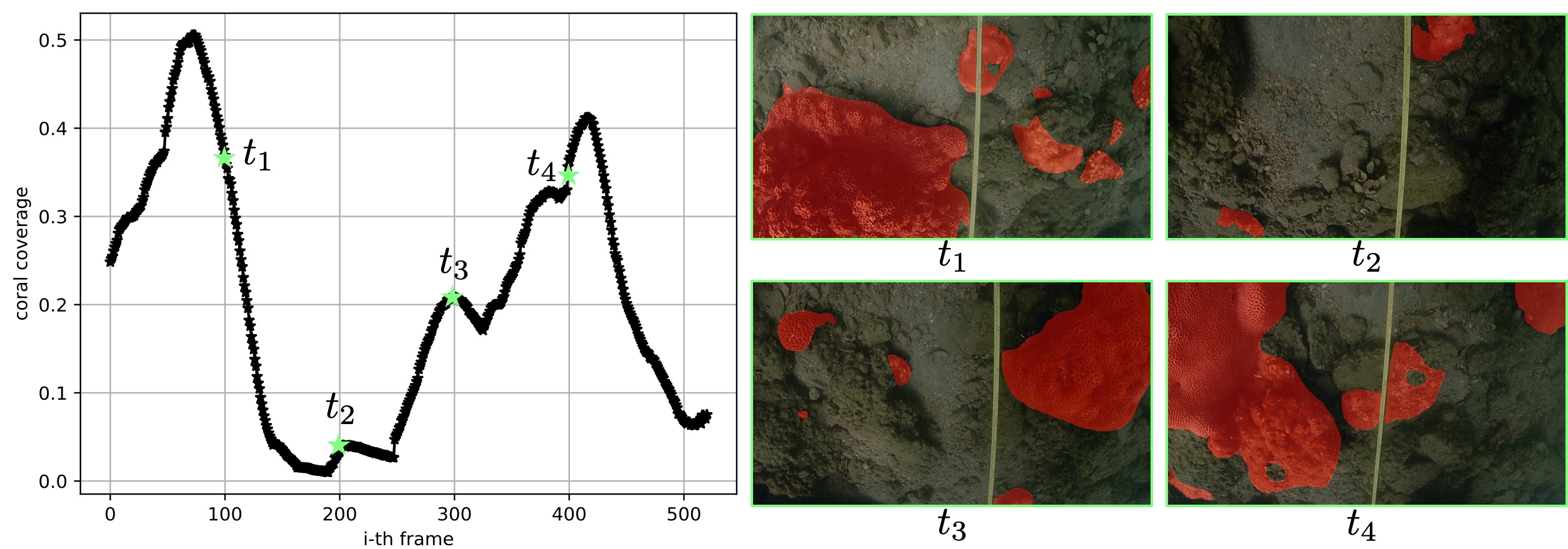}
  \end{center}
  \caption{Dense coral video segmentation for computing the coral coverage curve along the transect line. The segmentation results of images from some selected timestamps are provided for better illustration.}
  \label{fig:curve}
\end{figure}

Furthermore, the coral coverage curve along the whole transect line could also be computed by our method as demonstrated in Fig.~\ref{fig:curve}, which provides a more fine-grained and detailed spatial distribution of the coral reefs. As illustrated, we can easily observe the peak and shallow of the coral coverage for summarizing more sensitive findings. 

\subsection{Semantic 3D Reconstruction}
\begin{figure*}
  \begin{center}
    \includegraphics[width=\linewidth]{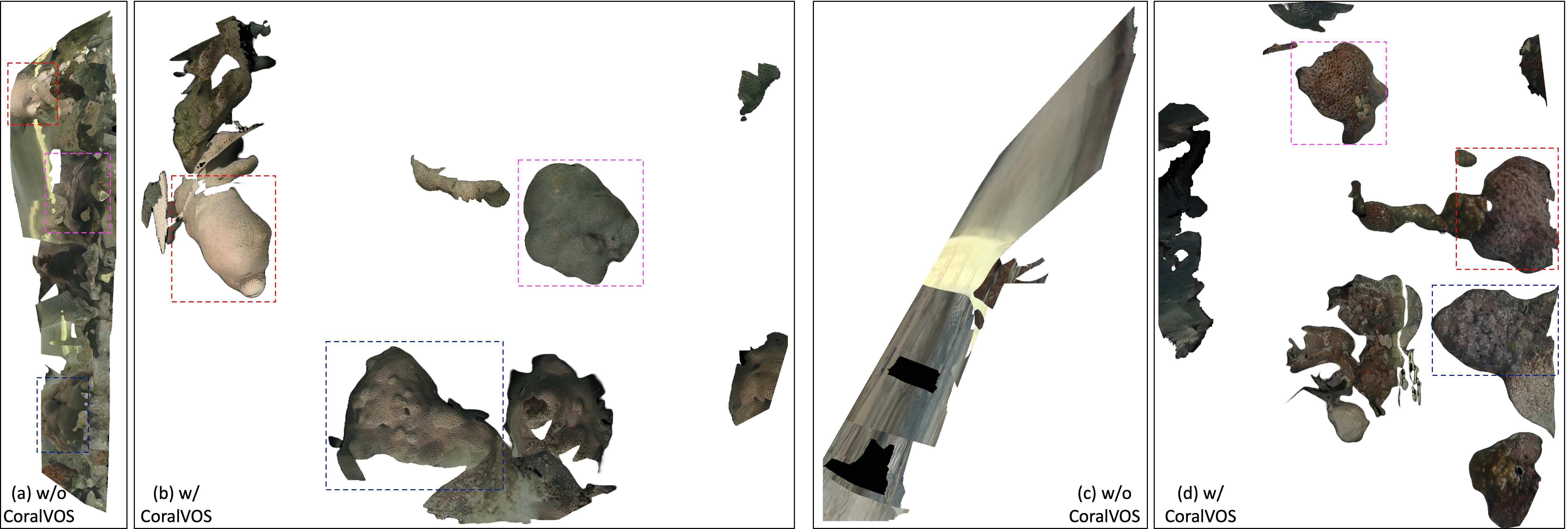}
  \end{center}
  \caption{The reconstructed 3D coral map under different settings. We observe that we could heavily promote 3D coral reconstruction performance and alleviate geometry distortions based on dense coral video segmentation. }
  \label{fig:3D_semantic}
\end{figure*}

The segmented coral masks in the 2D image space could be projected into the 3D space to promote the coral scene understanding in a 3D fashion. We perform 3D reconstruction based on structure-from-motion and obtain the reconstructed 3D model for better structure and geometry modeling of the coral colonies. Meanwhile, the generated coral masks by dense coral video segmentation are utilized as binary masks to remove the noisy background. We perform 3D reconstruction under ``original (w/o CoralVOS)'' and ``masked (w/ CoralVOS)'' settings and report the corresponding 3D reconstruction results in Fig.~\ref{fig:3D_semantic}. As demonstrated, our method could significantly reconstruct more \textbf{accurate}, \textbf{robust}, and \textbf{detailed} coral colonies without \textbf{geometry distortions}. Besides, we could also remove the background of the 3D model for better monitoring of the coral ecosystems. It is worth noting that we are performing dense 3D coral reconstruction, unlike the previous work~\cite{modasshir2020enhancing} that only modeled sparse points while not discriminating coral and non-coral regions. We also argue that VSLAM~\cite{campos2021orb} is highly subject to the efficiency and robustness of feature point detection algorithms~\cite{lowe2004distinctive,rublee2011orb}, the adverse underwater conditions will result in very few feature detection and matching. Combining the generated coral masks for promoting the feature point detection and matching performance will also lead to better reconstruction performance.

\subsection{Discussions and Limitation}
\noindent\textbf{Limitations}. Long-term video segmentation is much closer to practical applications. However, as the sequences in our CoralVOS dataset often span about 20 seconds, the performance of VOS models over long video sequences (\emph{e.g.}, minute-level) still needs to be explored. Bringing VOS into the long-term setting will increase demand for VOS models' re-detection capability. 

\noindent\textbf{Future work}. We could include the annotation of the coral status (\emph{e.g.}, healthy, half bleached, bleached and dead) into our dataset to help monitor the coral growth. The species-level annotations from coral experts could also be combined for more detailed and fine-grained coral reef analysis. We leave these as our future work.

\section{Conclusion}
By segmenting coral videos, researchers can effectively and efficiently identify and count coral coverage present in the footage. This supports biodiversity assessments and helps track changes in coral coverage over time. We propose a large-scale coral video segmentation dataset with densely labeled masks to promote the coral video segmentation performance. We have benchmarked various existing coral video segmentation algorithms on the proposed dataset and the experimental results demonstrate there is still a large room for coral video segmentation performance improvement. We also conduct an in-depth analysis and discuss the potential applications of our coral video segmentation. Our future work will address species-level coral video segmentation and monitor the coral status.
{
 \bibliographystyle{ieeetr}
\bibliography{icra}
}
\end{document}